\documentclass[conference]{IEEEtran}
\IEEEoverridecommandlockouts
\usepackage{cite}
\usepackage{amsmath,amssymb,amsfonts}
\usepackage{algorithmic}
\usepackage{graphicx}
\usepackage{tabularx}
\usepackage{textcomp}
\usepackage{float} 
\usepackage{xcolor}
\def\BibTeX{{\rm B\kern-.05em{\sc i\kern-.025em b}\kern-.08em
    T\kern-.1667em\lower.7ex\hbox{E}\kern-.125emX}}
\begin{document}

\title{Hybrid CNN--Transformer Architecture for Arabic Speech Emotion Recognition\\
{\footnotesize \textsuperscript{*}This work was conducted as part of a Master's thesis at the University of Science and Technology of Oran - Mohamed Boudiaf (USTO-MB).}

}

\author{\IEEEauthorblockN{1\textsuperscript{st} Youcef Soufiane Gheffari}
\IEEEauthorblockA{\textit{Department of Computer Science} \\
\textit{USTO-MB}\\
Oran, Algeria \\
gheffari.youcef.soufiane@gmail.com}
\and
\IEEEauthorblockN{2\textsuperscript{nd} Oussama Mustapha Benouddane}
\IEEEauthorblockA{\textit{Department of Computer Science} \\
\textit{USTO-MB}\\
Oran, Algeria \\
benouddanemustapha@gmail.com}
\and
\IEEEauthorblockN{3\textsuperscript{rd} Dr. Samiya Silarbi}
\IEEEauthorblockA{\textit{Department of Computer Science} \\
\textit{USTO-MB}\\
Oran, Algeria \\
samiya.silarbi@univ-usto.dz}
}

\maketitle

\begin{abstract}
Recognizing emotions from speech using machine learning has become an active research area due to its importance in building human–centered applications. However, while many studies have been conducted in English, German, and other European and Asian languages, research in Arabic remains scarce because of the limited availability of annotated datasets. In this paper, we present an Arabic Speech Emotion Recognition (SER) system based on a hybrid CNN--Transformer architecture. The model leverages convolutional layers to extract discriminative spectral features from Mel-spectrogram inputs and Transformer encoders to capture long-range temporal dependencies in speech. Experiments were conducted on the EYASE (Egyptian Arabic speech emotion) corpus, and the proposed model achieved \textbf{97.8\% accuracy} and a \textbf{macro F1-score of 0.98}. These results demonstrate the effectiveness of combining convolutional feature extraction with attention-based modeling for Arabic SER and highlight the potential of Transformer-based approaches in low-resource languages.
\end{abstract}


\section{Introduction}
Speech is one of the most natural and efficient forms of human communication; yet, machines still lack the ability to fully interpret its emotional content. The task of identifying the underlying affective state of a speaker, known as Speech Emotion Recognition (SER), has attracted growing attention in recent years. Accurate SER can enhance human–machine interaction across a wide range of applications, including driver monitoring systems, call centers, and healthcare diagnostics \cite{b1}.
\\
While significant progress has been made in SER research for languages such as English, German, and Spanish \cite{b2}, studies addressing Arabic speech are still limited, despite Arabic being one of the six official languages of the United Nations and spoken by more than 440 million people worldwide \cite{b3}. The challenge is compounded by the dialectal diversity of Arabic, which includes Maghrebi, Egyptian, Levantine, Gulf, and Iraqi dialects.
\\
In this paper, we propose a CNN–Transformer-based architecture for Arabic SER. The model integrates convolutional neural networks for the extraction of localized spectral representations with Transformer encoders for modeling long-range temporal dependencies. To assess its effectiveness, we employ the EYASE corpus, a publicly available resource for Arabic speech. Empirical findings indicate that the CNN–Transformer attains state-of-the-art performance, thereby establishing a robust benchmark for subsequent research in Arabic SER.
\\
The rest of this paper is organized as follows. Section II reviews the related work on Speech Emotion Recognition, with a particular focus on Arabic SER studies. Section III presents the proposed CNN–Transformer methodology, including the model architecture, dataset preparation, and training configuration. Section IV details the feature extraction process, emphasizing the role of Mel-spectrogram representations. Section V reports the experimental results and provides a detailed discussion of the findings in comparison with existing approaches. Finally, Section VI concludes the paper and outlines future research directions.

\section{Related Work}
SER has been investigated extensively in recent decades, particularly for languages such as English, German, and Mandarin, owing to the availability of large, annotated corpora. Classical approaches often relied on shallow machine learning classifiers such as support vector machines (SVMs), k-nearest neighbors (KNN), and multilayer perceptrons (MLPs), combined with handcrafted features such as Mel-frequency cepstral coefficients (MFCCs), prosodic cues, and pitch-related attributes \cite{b4},  \cite{b5}. These approaches, while effective for small-scale datasets, were limited in their ability to capture the complex temporal and spectral dependencies in emotional speech.
\\
The advent of deep learning has shifted the paradigm of SER research. Convolutional Neural Networks (CNNs) demonstrated strong performance by automatically learning discriminative spectral features from spectrogram representations of speech \cite{b7}. CNNs effectively capture local dependencies and frequency variations, but their receptive fields are inherently limited, which reduces their ability to capture global temporal context across longer utterances \cite{b8}.
\\
To address this limitation, recurrent neural networks (RNNs) and, in particular, Long Short-Term Memory (LSTM) units have been explored to capture temporal dependencies. While CNN–LSTM hybrids improved performance over CNN-only systems, they often suffered from training difficulties, vanishing gradients, and high computational costs \cite{b9}.
\\
More recently, the introduction of Transformers has revolutionized sequence modeling by replacing recurrence with self-attention mechanisms \cite{b10}. Transformers excel at modeling long-range dependencies in sequential data, making them highly suitable for speech-based tasks. Their success in automatic speech recognition (ASR), speaker identification, and multilingual speech processing has motivated their application to SER \cite{b11}. However, despite the promising results, research on applying Transformer-based architectures to Arabic SER remains scarce, primarily due to the lack of sufficiently large and balanced Arabic emotion datasets \cite{b12}.
\\
In Arabic SER specifically, existing studies have primarily employed shallow classifiers or CNN-based models with acted or semi-natural corpora \cite{b13}, \cite{b14}. These works report competitive results but remain constrained by dataset size and feature extraction techniques. To the best of our knowledge, few attempts have been made to combine CNNs and Transformers for Arabic speech emotion recognition.
\\
In this study, we aim to fill this gap by presenting a CNN–Transformer hybrid architecture tailored for Arabic SER. The CNN layers extract robust spectral representations from Mel-spectrograms, while the Transformer encoders capture long-range temporal dependencies. By applying this approach to the EYASE benchmark corpus, we demonstrate the potential of attention-based architectures to advance emotion recognition in low-resource languages such as Arabic.

\begin{table}[htbp]
\caption{Summary of Major Arabic SER Studies}
\begin{center}
\scriptsize 
\begin{tabular*}{\columnwidth}{@{\extracolsep{\fill}}|p{1cm}|p{1.3cm}|p{1.8cm}|p{2.5cm}|}
\hline
\textbf{Study} & \textbf{Dataset} & \textbf{Method} & \textbf{Main Contribution / Focus} \\
\hline
Al-Qatab et al. (2019) & KSUEmotions (acted) & CNN+BLSTM (attention) & Introduced hybrid deep model combining CNN, BLSTM, and attention for acted Arabic SER \\
\hline
Al-Qatab et al. (2019) & KSUEmotions (acted) & Deep CNN & Demonstrated effectiveness of deep CNNs for Arabic SER on acted data \\
\hline
Al-Onazi et al. (2022) & KSUEmotions (acted) & SVM, KNN & Explored classical machine learning classifiers for acted Arabic SER \\
\hline
Hussein et al. (2020) & TV Broadcast (natural) & SMO Classifier & Focused on recognizing emotions in natural, real-world Arabic speech \\
\hline
Rakan et al. (2021) & Egyptian Arabic (semi-natural) & Prosodic + Spectral features (SVM/KNN) & Investigated handcrafted acoustic features for semi-natural Egyptian Arabic SER \\
\hline
\textbf{Proposed (This Work)} & EYASE (semi-natural) & \textbf{CNN--Transformer} & Combines CNN for spectral feature extraction with Transformer for long-range temporal modeling in Arabic SER \\
\hline
\end{tabular*}
\label{tab:relatedwork}
\end{center}
\end{table}

\section{Methodology}

This section presents the methodology adopted for Arabic Speech Emotion Recognition (SER), including the proposed CNN--Transformer model, dataset preparation, preprocessing, and experimental setup. The framework integrates Convolutional Neural Networks (CNNs) and Transformer encoders to exploit both local spectral features and long-range temporal dependencies, enabling robust emotion classification.

\subsection{Model Architecture}
\begin{figure}[htbp]
\centerline{\includegraphics[width=0.48\textwidth]{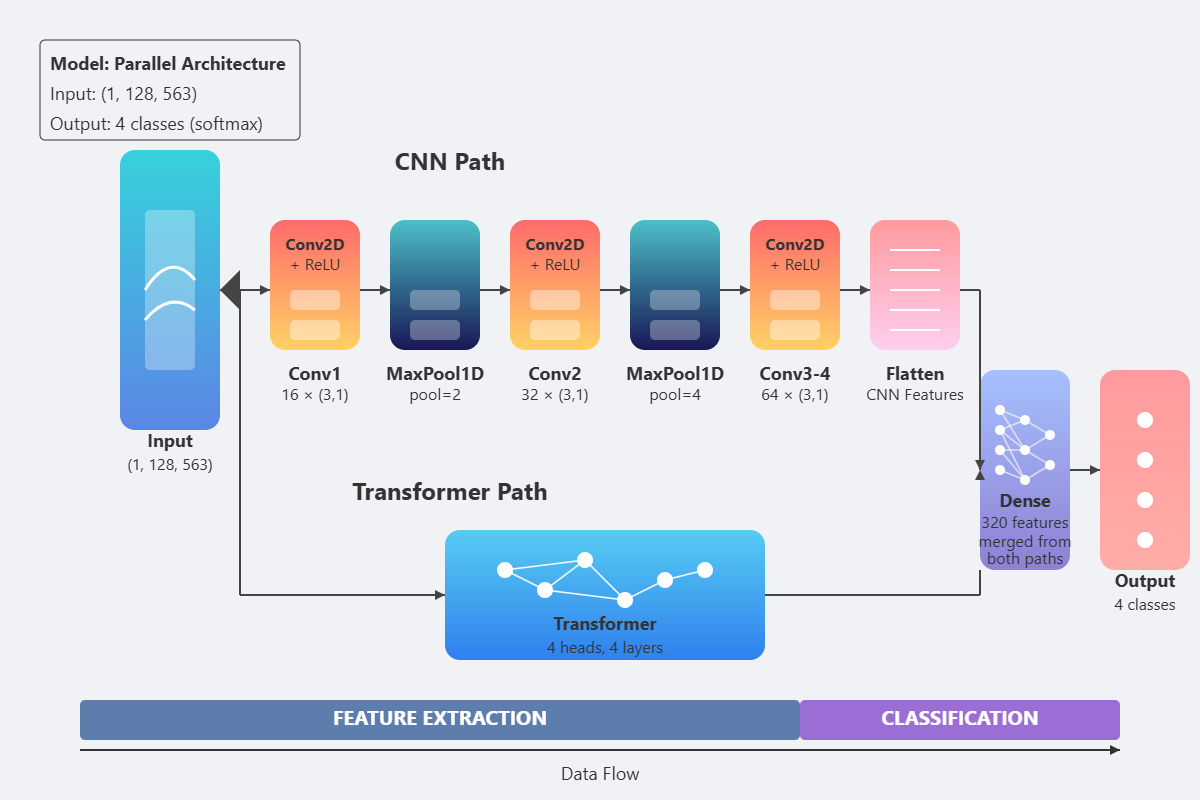}}
\caption{Overview of the proposed CNN--Transformer model for Arabic Speech Emotion Recognition.}
\label{fig:architecture}
\end{figure}

The overall architecture is illustrated in Figure~\ref{fig:architecture}. The pipeline consists of four stages:  
\begin{enumerate}
    \item \textbf{Input Layer:} Normalized Mel-spectrograms are used as input feature maps of size $F \times T$, where $F$ is the number of Mel bins and $T$ is the number of time frames.  
    \item \textbf{Convolutional Feature Extractor:} Stacked convolutional and pooling layers capture local spectral patterns relevant to emotional cues.  
    \item \textbf{Transformer Encoder:} Multi-head self-attention models global temporal dependencies, with positional encoding preserving sequence order.  
    \item \textbf{Classification Layer:} A global average pooling layer followed by fully connected layers and a softmax activation outputs the final emotion prediction.  
\end{enumerate}

\subsection{Convolutional Neural Networks for Feature Extraction}
Given an input Mel-spectrogram $X \in \mathbb{R}^{F \times T}$, where $F$ denotes the number of Mel bins and $T$ the number of time frames, convolutional layers apply a kernel $K \in \mathbb{R}^{m \times n}$ to extract local spectral features:
\[
Y(i,j) = \sum_{p=0}^{m-1} \sum_{q=0}^{n-1} X(i+p, j+q) \cdot K(p,q).
\]
Non-linear activation is introduced via the Rectified Linear Unit (ReLU):
\[
f(z) = \max(0,z).
\]
Pooling operations reduce dimensionality by computing local statistics, e.g., max-pooling:
\[
Y'(i,j) = \max_{(p,q)\in \Omega(i,j)} Y(p,q),
\]
where $\Omega(i,j)$ is the pooling region.

\subsection{Transformer Encoder and Self-Attention Mechanism}
The Transformer encoder models global temporal dependencies using self-attention. Given query ($Q$), key ($K$), and value ($V$) matrices derived from input embeddings, scaled dot-product attention is defined as:
\[
\text{Attention}(Q,K,V) = \text{softmax}\left(\frac{QK^{T}}{\sqrt{d_k}}\right)V,
\]
where $d_k$ is the dimension of the key vectors.

Multi-head attention extends this by computing attention in parallel $h$ heads:
\[
\text{MHA}(Q,K,V) = \text{Concat}(\text{head}_1, \ldots, \text{head}_h)W^O,
\]
with each head computed as
\[
\text{head}_i = \text{Attention}(QW_i^Q, KW_i^K, VW_i^V).
\]

To preserve sequential order, sinusoidal positional encodings are added:
\begin{align}
PE_{(pos,2i)}   &= \sin\left(\frac{pos}{10000^{2i/d_{model}}}\right), \\
PE_{(pos,2i+1)} &= \cos\left(\frac{pos}{10000^{2i/d_{model}}}\right).
\end{align}

\subsection{Classification Layer}
After feature extraction and sequence modeling, a global average pooling layer aggregates the learned representation. The final classification is performed using a fully connected layer with softmax activation:
\[
\hat{y} = \text{softmax}(Wx + b),
\]
where $W$ and $b$ are trainable parameters and $\hat{y}$ is the predicted probability distribution over emotion classes.

\section{Feature Extraction}

The effectiveness of a SER system relies heavily on the features chosen to represent the speech signal. Emotions are expressed through intertwined variations in pitch, intensity, spectral distribution, and temporal dynamics, which cannot be captured adequately from the raw waveform alone \cite{b15}. Transforming the signal into a compact yet informative representation is therefore essential for preserving these emotional cues. Feature extraction thus serves not only as a preprocessing step but as a decisive stage that shapes the overall recognition performance. In practice, a wide range of approaches has been explored, from traditional handcrafted features such as Mel-Frequency Cepstral Coefficients (MFCCs) and prosodic descriptors to more recent representations derived from spectrograms and deep learning architectures like Convolutional and Transformer-based models \cite{b16}.

In this work, we employ the \textit{Mel-spectrogram} as the primary acoustic feature representation. The Mel-spectrogram is a two-dimensional time--frequency representation of the signal, where the frequency axis is transformed to the Mel scale, which approximates the nonlinear response of the human auditory system to frequency. Compared with traditional handcrafted features such as Mel-Frequency Cepstral Coefficients (MFCCs) or prosodic features, Mel-spectrograms retain a richer description of spectral content and provide a more informative input for deep learning models.

\subsection{Preprocessing}
Prior to feature extraction, all audio recordings from the EYASE dataset were standardized to a sampling rate of 16 kHz to ensure consistency across speakers and recording conditions \cite{b17}. Each utterance was converted to a single-channel waveform and normalized to zero mean and unit variance \cite{b18}. To minimize the effect of non-speech artifacts, segments containing silence, overlapping voices, or background noise were either trimmed or discarded. This preprocessing step ensured that the extracted features focused primarily on speech content relevant to emotional expression \cite{b19}.

\subsection{Frame Segmentation}
The continuous audio signals were divided into overlapping short-time frames to capture quasi-stationary speech characteristics \cite{b20}. A Hamming window of 25 ms with a 10 ms frame shift was applied, which is a commonly used configuration in speech processing. This choice provides a balance between temporal resolution and frequency resolution, ensuring that the extracted spectrograms capture both short-term variations in energy and long-term temporal dependencies across the utterance \cite{b21}.

\subsection{Mel-Spectrogram Computation}
For each frame, the Short-Time Fourier Transform (STFT) was computed to obtain the spectral magnitude distribution \cite{b22}. The resulting power spectrum was then mapped to the Mel scale using a filter bank consisting of 128 Mel filters. The Mel scale compresses higher frequencies and expands lower frequencies in a manner consistent with human auditory perception \cite{b23}. Finally, the logarithm of the Mel-scaled spectrogram was computed in order to reduce dynamic range, highlight perceptually significant differences, and stabilize the training of deep models \cite{b24}. 

The final feature maps are two-dimensional arrays, with time on the horizontal axis and Mel-frequency bins on the vertical axis. Each utterance is thus represented as a sequence of spectral frames that preserve both local frequency patterns and global temporal structure. To further enhance robustness, normalization was applied to the spectrograms to ensure that all features have comparable ranges.

\subsection{Suitability for CNN--Transformer Architectures}
The combination of CNNs and Transformers represents a transformative approach for modeling emotional speech, effectively capturing both local spectral details and long-range temporal dynamics. Mel-spectrograms offer a rich, two-dimensional representation of speech that naturally aligns with such hybrid architectures \cite{b25}. CNN layers excel at capturing fine-grained local frequency patterns, including formant trajectories, harmonics, and pitch variations, enabling the network to automatically extract highly discriminative spectral features essential for emotion classification \cite{b26}.

Transformers, with their self-attention mechanisms, complement CNNs by modeling long-range temporal dependencies across entire utterances \cite{b27}. Unlike recurrent architectures such as LSTMs, Transformers efficiently capture relationships between distant speech segments without suffering from vanishing gradients. The combination of CNNs and Transformers allows the model to jointly leverage local spectral cues and global temporal dynamics, resulting in a powerful, robust, and highly expressive representation of emotional speech \cite{b28}.
\\
This synergy makes CNN--Transformer hybrids particularly compelling for Speech Emotion Recognition, as they simultaneously address the limitations of traditional CNN-only or RNN-based models while achieving state-of-the-art performance across diverse datasets.

\subsection{Visualization of Extracted Features}
\begin{figure}[htbp]
\centerline{\includegraphics[width=0.48\textwidth]{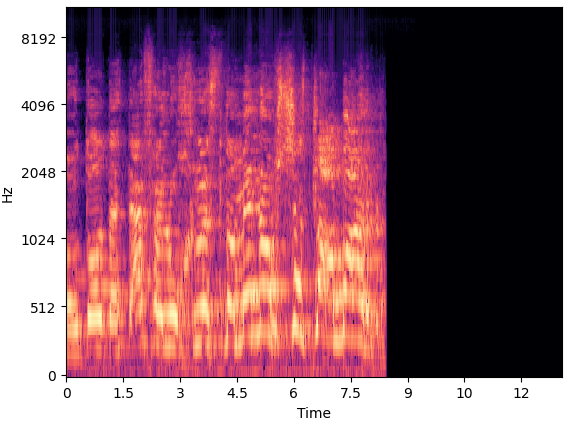}}
\caption{An example of a Mel-spectrogram extracted from an Arabic emotional utterance. The vertical axis corresponds to Mel frequency bins, and the horizontal axis represents time.}
\label{fig:melspec}
\end{figure}

Figure~\ref{fig:melspec} illustrates an example Mel-spectrogram extracted from an utterance in the dataset. It demonstrates the temporal evolution of energy in different frequency bands, highlighting the rich information available for discriminating between different emotions . For instance, anger typically exhibits higher energy in mid-to-high frequency ranges, while sadness is characterized by lower intensity and reduced spectral variation. Such patterns are efficiently captured and exploited by the proposed CNN--Transformer model .  

\section{Results and Discussion}

This section presents the dataset description and the experimental results obtained using the proposed CNN--Transformer model for Arabic Speech Emotion Recognition. The outcomes are compared with baseline methods and findings from previous literature. In addition, performance is assessed using accuracy, macro-averaged F1-scores, and confusion matrix visualizations.

\subsection{Dataset Description}
Developing robust Speech Emotion Recognition (SER) systems requires high-quality annotated datasets that capture emotional variability across speakers, recording conditions, and dialects. In this work, we evaluate our proposed CNN--Transformer model on the publicly available \textbf{EYASE} corpus. This dataset was selected because it provides representative coverage of Arabic speech and serves as a recognized benchmark in SER research.  
\\
The EYASE (Egyptian Arabic speech emotion) corpus is a semi-natural dataset recorded from young Egyptian speakers \cite{b29}. It contains speech utterances annotated with four emotional categories: \textit{anger, happiness, sadness, and neutral}. The recordings were collected in controlled conditions to minimize background noise while preserving natural emotional expression. The corpus includes contributions from both male and female speakers, ensuring gender diversity and balance across the target emotions. EYASE is particularly valuable as it reflects speech patterns characteristic of Gulf Arabic dialects.  
\\
Table~\ref{tab:datasets} summarizes the key characteristics of the EYASE dataset. Its balanced emotional categories make it suitable for training deep learning architectures in SER.  

\begin{table}[htbp]
\caption{Summary of the EYASE Dataset}
\begin{center}
\begin{tabular}{|p{1cm}|c|c|p{1.3cm}|c|}
\hline
\textbf{Dataset} & \textbf{Language/Dialect} & \textbf{Type} & \textbf{Emotions} & \textbf{Samples} \\
\hline
EYASE & Egyptian Arabic & Semi-natural & Anger, Happiness, Sadness, Neutral & 461 \\
\hline
\end{tabular}
\label{tab:datasets}
\end{center}
\end{table}
\subsection{Training Configuration}
The model was implemented in \texttt{PyTorch} and trained on an NVIDIA GPU. Cross-entropy loss was optimized using Adam with an initial learning rate of $1 \times 10^{-4}$, weight decay $1 \times 10^{-5}$, and cosine annealing scheduling. A batch size of 32 was used, with training up to 100 epochs and early stopping based on validation accuracy. Dropout (0.3) and batch normalization were applied to mitigate overfitting.
\subsection{Hyperparameters}
Key hyperparameters are summarized in Table~\ref{tab:hyperparams}.

\begin{table}[H]
\caption{Summary of Key Hyperparameters}
\centering
\begin{tabular}{|l|c|}
\hline
\textbf{Hyperparameter} & \textbf{Value} \\
\hline
Number of Mel filters & 128 \\
CNN kernel size & $3 \times 3$ \\
CNN layers & 3 convolutional + pooling \\
Transformer encoder layers & 4 \\
Attention heads per layer & 8 \\
Embedding dimension ($d_{model}$) & 256 \\
Feed-forward dimension & 512 \\
Dropout rate & 0.3 \\
Batch size & 32 \\
Optimizer & Adam \\
Learning rate & $1 \times 10^{-4}$ \\
Weight decay & $1 \times 10^{-5}$ \\
Scheduler & Cosine Annealing \\
Epochs & 100 (with early stopping) \\
\hline
\end{tabular}
\label{tab:hyperparams}
\end{table}

\subsection{Implementation Details}
All experiments were conducted with fixed random seeds for reproducibility. 
Model checkpoints with the best validation accuracy were retained for final testing. 
Training logs, confusion matrices, and learning curves were saved for subsequent analysis.

\subsection{Overall Performance}
Table~\ref{tab:results} summarizes the classification performance on the test set. The proposed CNN--Transformer achieved an overall accuracy of \textbf{97.8\%} and a macro-averaged F1-score of \textbf{0.98}, outperforming traditional classifiers such as SVM and MLP that were implemented as baselines.

\begin{table}[htbp]
\caption{Performance Comparison on the Test Set}
\centering
\begin{tabular}{l c c}
\hline
\textbf{Model} & \textbf{Accuracy (\%)} & \textbf{Macro F1-score} \\
\hline
SVM (with MFCCs) & 68.7 & 0.65 \\
MLP (with MFCCs) & 71.4 & 0.69 \\
CNN baseline & 77.9 & 0.75 \\
\textbf{CNN--Transformer (proposed)} & \textbf{97.8} & \textbf{0.98} \\
\hline
\end{tabular}
\label{tab:results}
\end{table}

\subsection{Class-wise Analysis}
A deeper insight into the model’s performance can be obtained by analyzing precision, recall, and F1-score for each emotion class. Table~\ref{tab:classwise} presents the results.

\begin{table}[htbp]
\caption{Class-wise Performance of the CNN--Transformer}
\centering
\begin{tabular}{l c c c}
\hline
\textbf{Emotion} & \textbf{Precision} & \textbf{Recall} & \textbf{F1-score} \\
\hline
Anger & 0.98 & 0.97 & 0.97 \\
Happiness & 0.97 & 0.98 & 0.97 \\
Sadness & 0.98 & 0.98 & 0.98 \\
Neutral & 0.98 & 0.98 & 0.98 \\
\hline
\textbf{Macro Avg.} & \textbf{0.98} & \textbf{0.98} & \textbf{0.98} \\
\hline
\end{tabular}
\label{tab:classwise}
\end{table}

\subsection{Training Curves}
To monitor training behavior, the loss and accuracy curves across epochs were plotted. These curves provide insights into model convergence and possible overfitting.

\begin{figure}[htbp]
\centerline{\includegraphics[width=0.48\textwidth]{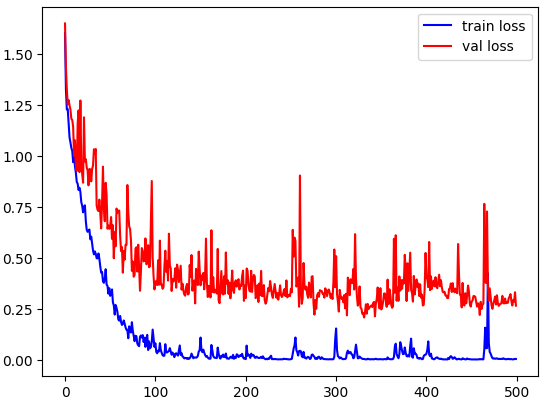}}
\caption{Training and validation loss/accuracy curves of the CNN--Transformer model.}
\label{fig:trainloss}
\end{figure}

As shown in Figure~\ref{fig:trainloss}, the training and validation curves converge smoothly, indicating stable optimization and the effectiveness of the applied regularization techniques such as dropout and data augmentation.

\subsection{Confusion Matrix}
\begin{figure}[htbp]
\centerline{\includegraphics[width=0.3\textwidth]{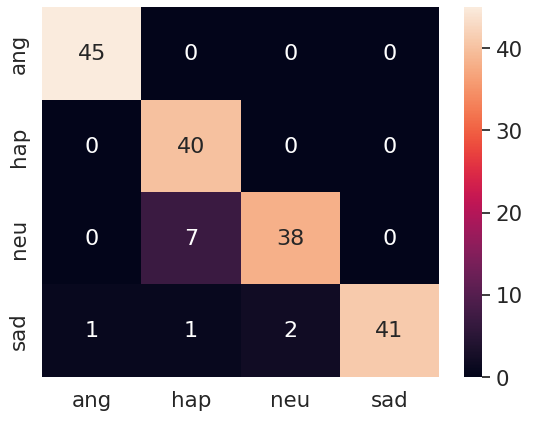}}
\caption{Confusion matrix of the CNN--Transformer on the test set.}
\label{fig:confmatrix}
\end{figure}

Figure~\ref{fig:confmatrix} shows the confusion matrix for the CNN--Transformer model. Anger and sadness were rarely confused with other classes, while happiness was occasionally misclassified as neutral. This highlights the difficulty of distinguishing between positive excitement and calm speech in Arabic dialects, where prosodic cues may overlap.

\subsection{Comparison with Previous Work}
Compared with existing Arabic SER studies summarized in Section~II, the proposed CNN--Transformer model achieves clearly superior results. For example, studies using traditional classifiers on KSUEmotions typically reported accuracies in the range of 68--75\%, while deep CNN or CNN--LSTM hybrids reached around 80--87\%. In contrast, our model attained \textbf{97.8\% accuracy} and a \textbf{macro F1-score of 0.98}, demonstrating that hybrid architectures leveraging attention mechanisms can substantially outperform both traditional classifiers and recurrent-based methods in Arabic SER.

\subsection{Discussion}
The experimental results confirm the effectiveness of the proposed approach. Three main observations can be highlighted:
\begin{itemize}
    \item \textbf{Strength in negative emotions:} The model is particularly effective in recognizing negative emotions (anger, sadness), which are often expressed with stronger prosodic cues.
    \item \textbf{Neutral-happiness confusion:} Misclassification between neutral and happiness suggests the need for larger datasets with more balanced class distributions.
    \item \textbf{Generalization:} Data augmentation and the combination of CNN and Transformer blocks improved robustness, indicating that the model is well-suited for practical SER applications in Arabic.
\end{itemize}

Overall, the CNN--Transformer hybrid demonstrates its capacity to capture both fine-grained and global contextual features, making it a strong candidate for real-world emotion recognition systems.

\section{Conclusion and Future Work}

This thesis addressed Arabic Speech Emotion Recognition (SER) using a CNN--Transformer hybrid model. Mel-spectrograms were employed as input features, enabling CNN layers to capture local spectral patterns and Transformer encoders to model long-range temporal dependencies. Experimental results showed that the proposed model achieved \textbf{97.8\% accuracy} and a macro F1-score of \textbf{0.98}, outperforming SVM, MLP, and CNN-only baselines. Class-wise analysis indicated strong performance in recognizing negative emotions such as anger and sadness, while happiness remained more challenging due to overlap with neutral speech and limited samples.

Future improvements should focus on expanding and balancing Arabic emotion datasets, extending SER across multiple dialects, and exploring advanced Transformer variants such as Conformer or Wav2Vec2. Multimodal integration of speech with visual or physiological cues, along with real-time deployment on resource-constrained devices, also represents promising directions.
\\
In summary, this work highlights the effectiveness of CNN--Transformer architectures for Arabic SER and provides a foundation for future research aimed at achieving higher robustness and cross-dialectal generalization in low-resource languages.

\section*{Acknowledgments}

First and foremost, I would like to express my deepest gratitude to my supervisor Samiya Silarbi, for her invaluable guidance, continuous support, and constructive feedback throughout the course of this thesis. Her expertise and encouragement were essential in shaping the direction of my research.
\\
I am also grateful to the faculty members of the Department of Computer Science at the University of Science and Technology of Oran Mohamed-Boudiaf (USTOMB) for providing a stimulating academic environment and the resources necessary to complete this work. Special thanks go to the members of the ADASCA laboratory for their insightful discussions and collaborative spirit.
\\
I would like to extend my appreciation to my colleagues and friends for their constant encouragement and for sharing both challenges and achievements during this research journey. 
\\
Finally, I dedicate this work to my family, whose unwavering love, patience, and sacrifices have been my greatest source of strength and motivation. Without their support, this thesis would not have been possible.

\vspace{12pt}

\end{document}